\documentclass[letterpaper, 10 pt, conference]{ieeeconf}  

\IEEEoverridecommandlockouts                              

\usepackage{amsmath} 
\usepackage{amssymb}  

\usepackage{amsmath,amsfonts,bm}

\def\eqref#1{equation~\ref{#1}}

\def\1{\bm{1}}

\DeclareMathAlphabet{\mathsfit}{\encodingdefault}{\sfdefault}{m}{sl}
\SetMathAlphabet{\mathsfit}{bold}{\encodingdefault}{\sfdefault}{bx}{n}

\DeclareMathOperator*{\argmin}{arg\,min}

\usepackage{url}
\usepackage{graphicx}
\usepackage{booktabs}
\usepackage{multirow}
\usepackage{array}
\usepackage{lscape}
\usepackage{longtable}
\usepackage{subcaption}
\usepackage{wrapfig}    
\usepackage{tabularx}
\usepackage{siunitx}
\graphicspath{images/}
\usepackage{subfiles}
\usepackage{xcolor}
\usepackage[normalem]{ulem}
\makeatletter
\let\NAT@parse\undefined
\makeatother
\usepackage[colorlinks=true, linkcolor=black, citecolor=blue]{hyperref}
\newcolumntype{x}[1]{>{\centering\let\newline\\\arraybackslash\hspace{0pt}}p{#1}}

\title{\LARGE \bf
Extraneousness-Aware Imitation Learning
}

\author{Ray Chen Zheng*$^{1, 4}$, Kaizhe Hu*$^{1, 4}$, Zhecheng Yuan$^{1, 4}$, Boyuan Chen$^{3}$, Huazhe Xu$^{1, 2, 4}$
\thanks{*Denotes equal contribution.}
  \thanks{$^{1}$Tsinghua University,\ $^{2}$Shanghai AI Lab,\ $^{3}$Massachusetts Institute of Technology,\  $^{4}$Shanghai Qi Zhi Institute}
  \thanks{{Contact: {\tt\small \href{mailto:zhengrc19@mails.tsinghua.edu.cn}{zhengrc19@mails.tsinghua.edu.cn}}.}}
}

\begin{document}

\maketitle
\thispagestyle{empty}
\pagestyle{empty}

\maketitle

\begin{abstract}
Visual imitation learning provides an effective framework to learn skills from demonstrations.
However, the quality of the provided demonstrations usually significantly affects the ability of an agent to acquire desired skills. 
Therefore, the standard visual imitation learning assumes near-optimal demonstrations, which are expensive or sometimes prohibitive to collect. 
Previous works propose to learn from \textit{noisy} demonstrations; however, the noise is usually assumed to follow a context-independent distribution such as a uniform or gaussian distribution. 
In this paper, we consider another crucial yet underexplored setting --- imitation learning with task-irrelevant yet locally consistent segments in the demonstrations (e.g., wiping sweat while cutting potatoes in a cooking tutorial). 
We argue that such noise is common in real world data and term them as ``extraneous'' segments. 
To tackle this problem, we introduce Extraneousness-Aware Imitation Learning (EIL), a self-supervised approach that learns visuomotor policies from third-person demonstrations with extraneous subsequences. 
EIL learns action-conditioned observation embeddings in a self-supervised manner and retrieves task-relevant observations across visual demonstrations while excluding the extraneous ones. 
Experimental results show that EIL outperforms strong baselines and achieves comparable policies to those trained with perfect demonstration on both simulated and real-world robot control tasks.
The project page can be found here: \url{https://sites.google.com/view/eil-website}. 

\end{abstract}

\begin{figure*}[t]
    \centering
    \includegraphics[width=0.9\textwidth]{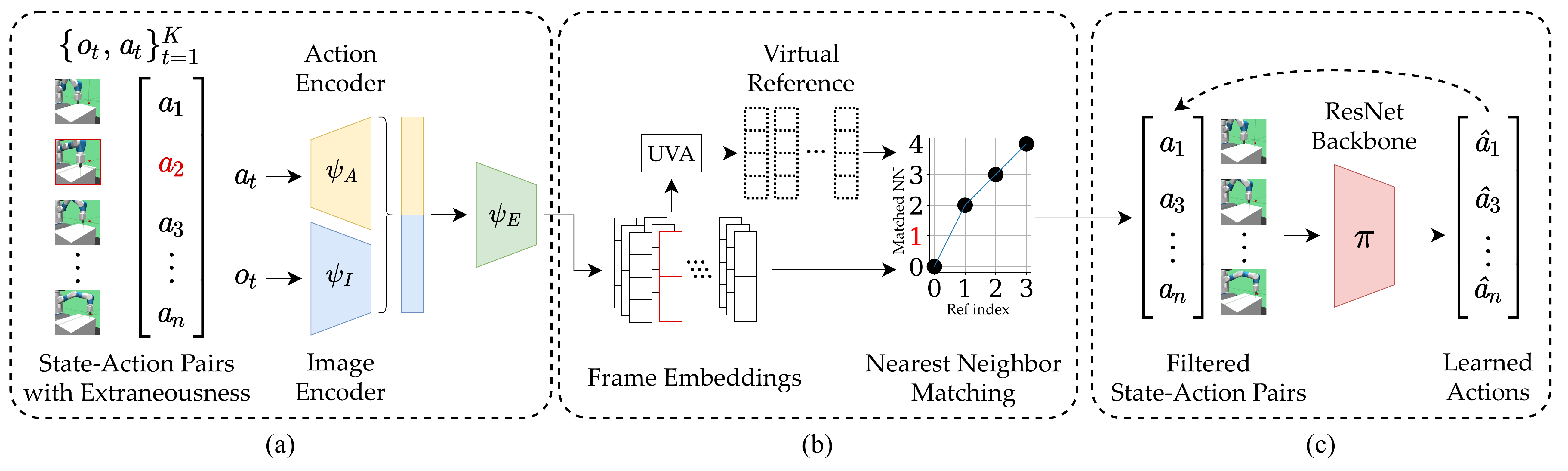}
    \caption{\textbf{Extraneousness-Aware Imitation Learning (EIL). } The overall framework contains 3 components: (a) It encodes the state-action pairs into representation. (b) It takes in the embeddings and process them with unsupervised voting-based alignment (UVA) algorithms. (c) It performs visual imitation learning with the aligned state action pairs. We note that (b) can be a simple filtering algorithm when reference trajectories are available. }
    \label{fig:eil}
\end{figure*}

\section{Introduction}

Imitation learning (IL) enables intelligent agents to acquire various skills from demonstrations~\cite{argall2009survey, schaal1997learning}; recent advances also extend IL to the visual domain~\cite{pathak2018zero, zhang2018deep, young2020visual, zhu2018reinforcement}. However, in contrast to how humans learn from demonstrations, artificial agents usually require ``clean'' data sampled from expert policies. Some recent literatures~\cite{tangkaratt2020robust, brown2019extrapolating, wu2019imitation, tangkaratt2020variational} propose methods to perform imitation learning from noisy demonstrations. However, many of these methods are state-based, and are limited by their requirements such as additional labels or assumptions about the noise. Despite the effort, real-world data may contain extraneous segments that can hardly be defined or labelled. For example, when learning to cut potatoes from videos, humans can naturally ignore some of the demonstrators’ extraneous actions like wiping sweat in the halfway. This distinction raises the question naturally: how can we leverage the rich range of unannotated visual demonstrations for imitation learning without being hindered by their noise?

In this paper, we propose Extraneousness-Aware Imitation Learning (EIL) that enables agents to imitate from noisy video demonstrations with extraneous segments. Our method allows agents to identify extraneous subsequences via self-supervised learning and selectively perform imitation from the task-relevant parts. Specifically, we train an action-conditioned encoder through temporal cycle-consistency~(TCC) loss~\cite{dwibedi2019temporal} to obtain the embeddings of each observation. In this way, the observations of similar progress in the demonstrations will gain similar embeddings. Then, we propose an Unsupervised Voting-based Alignment algorithm~(UVA) to filter task-irrelevant frames across video clips. Finally, we introduce a few tasks to benchmark the performance of imitation learning from noisy data with extraneous sequences. 

We evaluate our method on multiple visual-input robot control tasks in both simulator and the real-world. The experiment results suggest that the proposed encoder can produce embeddings useful for extraneousness detection. As a result, EIL outperforms various baselines and achieves comparable performance to those trained with perfect demonstrations. 

Our contributions can be summarized as follows: 
1) We propose a meaningful yet underexplored setting of visual imitation learning from demonstrations with extraneous segments
2) We introduce Extraneousness-Aware Imitation Learning (EIL) that learns selectively from the task-relevant parts by leveraging action-conditioned embeddings and alignment algorithms.
3) We introduce datasets with extraneous segments over several simulated or real-world tasks and demonstrate our method's empirical effectiveness.

\section{Related Works}
\subsection{Learning from Noisy Demonstration}
Imitation learning~\cite{argall2009survey, schaal1997learning, ho2016generative, dadashi2020primal} includes behavior cloning~\cite{bain1995framework, pomerleau1991efficient} which aims to copy the behaviors from the demonstration, and inverse reinforcement learning~\cite{ng2000algorithms} that infers the reward function for learning policies. However, these methods usually assume access to expert demonstrations, which are hard to obtain in practice.

Recent works try to tackle the imitation learning problem when the demonstrations are noisy. However, through this line of research~\cite{kaiser1995obtaining, grollman2012robot, kim2013learning, burchfiel2016distance, brown2019extrapolating, wu2019imitation, tangkaratt2020variational, sasaki2020behavioral}, the vast majority of the works are done in the low-dimensional state space rather than the high-dimensional image space. Furthermore, it is common in previous works~\cite{sasaki2020behavioral} to assume the noise is sampled from an \textit{a priori} distribution. Methods designed specifically for such noise might fail completely when the noise violates the assumption. Recently, more attentions are drawn to learning from realistic visual demonstrations, e.g., Chen et al. propose to learn policies from ``in-the-wild'' videos~\cite{chen2021learning}. While the method achieves impressive results, they focus on dealing with diverse demonstrations without considering the ``extraneousness'' explicitly.

\subsection{Self-Supervised Learning from Videos and its Application \\\indent to Control and Robotics Tasks}
Self-supervised learning~(SSL) from videos can learn visual representations with temporal information for different downstream tasks from unlabeled data~\cite{gordon2020watching, pathak2017learning, srivastava2015unsupervised, mathieu2015deep, jayaraman2015learning, agrawal2015learning, goroshin2015unsupervised}. A recent line of research utilizes SSL for learning correspondences~\cite{jabri2020space, wang2019learning, vondrick2018tracking, dwibedi2019temporal, hadji2021representation, purushwalkam2020aligning}. Specifically, Dwibedi et al. propose to find correspondences across time in multiple videos with the help of cycle-consistency, where frames with similar progress will be encoded to similar embeddings~\cite{dwibedi2019temporal, zhou2016learning, zhu2017unpaired}. These methods offer a welcoming way to leverage unlabeled and noisy real-world data. In recent years, SSL also promises to help with visuomotor tasks in control and robotics~\cite{nair2022r3m, xiao2022masked}. For example, TCN~\cite{sermanet2018time} learns a self-supervised temporal-consistent embedding for imitation learning and reinforcement learning. XIRL~\cite{zakka2021xirl} learns a self-supervised embedding that estimates task progress for inverse reinforcement learning. \cite{zhang2020learning, smith2019avid, xiong2021learning} directly map the observations such as images to the target domain. The distinction between EIL and previous works is that we tackle the problem where demonstrations have extraneous subsequences, rather than different visual appearances, view points, or embodiments.

\section{Method}

\begin{figure*}[thbp]
\begin{center}
\includegraphics[width=0.9\textwidth]{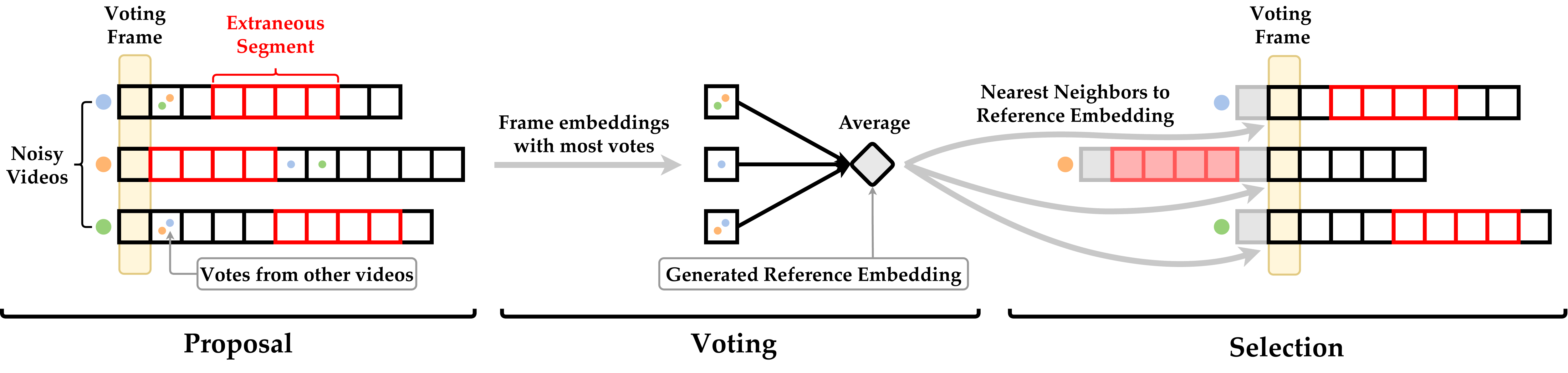}
\end{center}
\caption{\textbf{Unsupervised Voting-based Alignment (UVA).} The three-stage structure includes: 1) Proposal where each frame performs nearest neighbor voting according to their embeddings. 2) Voting where all the selected frame embeddings are averaged out to get a virtual reference embedding. 3) Selection where we select the actual frame as our training data based on the estimated embedding.}
\label{fig:uva}
\end{figure*}

In this section, we first describe the problem setup and then introduce Extraneousness-Aware Imitation Learning~(EIL), a simple yet effective approach for learning visuomotor policies from videos that have extraneous subsequences.
\subsection{Problem Statement}
We consider the setting where an agent aims at learning visuomotor policies from $K$ video demonstrations $\{\mathcal{D}_i\}_{i=1}^K$. In the $i$\textsuperscript{th} video, the $j$\textsuperscript{th} observation $o_j^i$ is paired up with its corresponding action $a_j^i$. For each sequence in the demonstration set, there are $L$ extraneous subsequences $\{\mathcal{E}_n\}_{n=1}^L$ that are task-irrelevant yet locally consistent. In contrast to existing works that have various assumptions about the noise, our setting only assumes each video to contain more than 50\% of task-relevant content~\cite{angluin1988learning, natarajan2013learning}. 

The imitation agent takes a high-dimensional observation $o_t$ as input and outputs an action $\hat{a}_t$ at timestep $t$. To successfully imitate from the aforementioned demonstrations, the agent needs to reason about what the task-relevant parts are and rule out the extraneous subsequences. 

\subsection{Extraneousness-Aware Imitation Learning~(EIL)}
\subsubsection{Overview} EIL is a general framework for imitating from videos with extraneous subsequences.
The intuition of EIL is that the task-relevant parts among different demonstrations will share similar semantics meaning in the latent space, thus they can be aligned with each other. Following such intuition, when more than one demonstration sequence are given, we can match their embeddings to retrieve the task-relevant parts. In the case where a perfect reference demonstration is available, we can match frames in other sequences with that of the reference trajectory. However, in most cases, such a reference is hard to obtain. Hence, we propose an unsupervised alignment algorithm to retrieve task-relevant parts from a set of noisy demonstrations. 

Figure \ref{fig:eil} demonstrates the overview of EIL. In Figure \ref{fig:eil}(a), we learn a temporal representation of each frame conditioned on both its visual observation and action through temporal cycle consistency loss. After obtaining the representation, as shown in Figure \ref{fig:eil}(b), we propose an unsupervised voting method to perform video filtering when no perfect demonstration is available. Finally, as described in Figure \ref{fig:eil}(c), we perform standard visual imitation learning on top of the denoised data from the alignment procedure.

\subsubsection{Action-conditioned Temporal Cycle Consistency Representation Learning}
We first learn representations that encode temporal information for frame alignment across different video demonstrations. We train an image encoder $\psi_{I}$ and an action encoder $\psi_{A}$ that embed the observations and actions into corresponding features $\psi_{I}(o)$ and $\psi_{A}(a)$.
Then, we concatenate $\psi_{I}(o)$ and $\psi_{A}(a)$ to a multi-layer perceptron (MLP) $\psi_{E}$ to obtain the embeddings that have temporal correspondences between two sequences. For simplicity, we use two demonstration sequences $S$ and $T$ and their computed embeddings $U=\{u_1, u_2, \cdots, u_{N}\}$ and $V=\{v_1, v_2, \cdots, v_M\}$ as an example. $N$ and $M$ denotes the sequence lengths respectively. 

The main goal here is to encourage cycle-consistency between the two embedding sequences. For any $u_i \in U$, we find the nearest neighbor, $v_j = \argmin_{v \in V} ||v - u_i||$. Then we repeat the procedure and find $u_k = \argmin_{u \in U} ||v_j - u||$ which is the nearest neighbor for $v_j$. When $i = k$, the embedding $u_i$ is cycle-consistent. To optimize the cycle-consistency, we use a differentiable matching loss: for the selected $u_i$, we compute the soft nearest neighbor by
$\displaystyle \tilde{v} = \sum_{j}^M \alpha_j v_j, \text{where } \alpha_j = \frac{\exp{(-|| u_i -v_j||^2)}}{\sum_k^M \exp{(-||u_i - v_k||^2)}}$
Then, we compute the ``cycle-back'' soft nearest neighbor to $\tilde{v}$ similarly:
$\displaystyle \tilde{u} = \sum_{k}^N \beta_k u_k, \text{where } \beta_k = \frac{\exp{(-|| \tilde{v} - u_k||^2)}}{\sum_j^N \exp{(-||\tilde{v}-u_j||^2)}}$
The predicted index $\hat{i}$ is calculated by $\hat{i}=\sum_k^{N} \beta_k k$. To obtain the cycle-consistency, $\hat{i}$ should be close to the true $i$. Hence, we minimize the loss with an imposed Gaussian prior and variance regularization \cite{dwibedi2019temporal}

\begin{equation}
    \mathcal{L} = \frac{|i - \hat i|^2}{\sigma^2} + \lambda \log(\sigma)
\end{equation}
where $\sigma = \sum_k^N \beta_k (k-\hat i)^2$ and $\lambda$ is the regularization weight. At test time, the indices can be rounded to integers.

\subsubsection{Unsupervised Voting-based Frame Alignment (UVA)} After obtaining the embedding for each observation-action pair, we try to align frames and drop extraneous frames according to a frame-wise similarity in the latent space. 

To achieve this objective, we propose a voting-based frame matching algorithm that can remove the extraneous segments from a set of videos. A conceptual illustration is shown in Figure  \ref{fig:uva}.
For $K$ demonstration sequences, we initially mark the first frame of each video as the ``voting frame''. The distance and nearest neighbor mentioned below is in the embedding space. Our algorithm can be described as below: 
\begin{itemize}
    \item \textbf{Proposal.} For each video, find nearest neighbor to the ``voting frame'' among $K-1$ other videos.
    \item \textbf{Voting.} In each video, the frame selected as nearest neighbor the most times is marked as a new ``voting frame`` of that video. We average the embeddings of all the newly selected ``voting frames'' to get a virtual reference embedding representing the current progress.
    \item \textbf{Selection.} We select the nearest neighbor of the virtual embedding in each video as the new ``voting frame''. A causal restriction to select only in frames after the current ``voting frame'' is applied.
\end{itemize}

In a simpler setting where we have access to a perfect reference demonstration, our algorithm degenerates to simply picking frames in each video that are nearest neighbors to each frame of the reference. 

\subsubsection{Visual Imitation Learning} As shown in Figure~\ref{fig:eil}(c),  we perform standard visual imitation learning to learn a policy $\pi$ that minimizes distance between the predicted actions and the ground-truth actions using the state-action pairs that are selected previously. Specifically, for continuous actions, we use the $\ell_2$ loss: $L=||a_i - \widehat{a_i}||^2$, where $a_i$ and $\widehat{a_i}$ are the ground truth and predicted action respectively.
For discrete actions, we use cross-entropy loss instead. We use ResNet-18 as our policy network to process the image input.

\section{Experiment}

\begin{table*}[h]
\centering
\caption{Averaged success rates and standard deviations of EIL and other baselines for simulated control tasks.}
\label{tab:success-rate}
\resizebox{0.8\textwidth}{!}{%
\begin{tabular}{c|c|cccccc}
\toprule
\textbf{Task} & \textbf{Oracle} & \textbf{EIL (Ours)} & \textbf{Behavior Cloning} & \textbf{RL} & \textbf{TCN} & \textbf{RIL-Co} & \textbf{Random Policy} \\ \midrule
\textit{Reach} & 100\% & \textbf{85\% $\pm$ 3\%} & 77\% $\pm$ 4\% & 20\% $\pm$ 5\% & 80\% $\pm$ 3\% & 82\% $\pm$ 5\% & 4\% $\pm$ 3\% \\
\textit{Push} & 100\% & \textbf{79\% $\pm$ 3\%} & 66\% $\pm$ 4\% & 16\% $\pm$ 5\% & 76\% $\pm$ 3\% & 22\% $\pm$ 6\% & 0\% \\
\textit{Stir} & 100\% & \textbf{83\% $\pm$ 3\%} & 55\% $\pm$ 4\% & 32\% $\pm$ 7\% & 61\% $\pm$ 4\% & 54\% $\pm$ 7\% & 0\% \\ \bottomrule
\end{tabular}%
}
\end{table*}

\begin{table*}[h]
\centering
\caption{Averaged minimum distances and standard deviations of EIL and other baselines for simulated control tasks.}
\label{tab:min-dist}
\resizebox{\textwidth}{!}{%
\begin{tabular}{c|c|cccccc}
\toprule
\textbf{Task} & \textbf{Oracle} & \textbf{EIL (Ours)} & \textbf{Behavior Cloning} & \textbf{RL} & \textbf{TCN} & \textbf{RIL-Co} & \textbf{Random Policy} \\ \midrule
\textit{Reach} & 0.0175 $\pm$ 0.0105 & \textbf{0.0339 $\pm$ 0.0166} & 0.0374 $\pm$ 0.0174 & 0.0716 $\pm$ 0.0224 & 0.0371 $\pm$ 0.0172 & 0.0371 $\pm$ 0.0153 & 0.2059 $\pm$ 0.0675 \\
\textit{Push} & 0.0242 $\pm$ 0.0060 & \textbf{0.0290 $\pm$ 0.0350} & 0.0403 $\pm$ 0.0435 & 0.1768 $\pm$ 0.1790 & 0.0305 $\pm$ 0.0275 & 0.1011 $\pm$ 0.0484 & 0.1456 $\pm$ 0.0152 \\
\textit{Stir} & 0.0108 $\pm$ 0.0017 & \textbf{0.0388 $\pm$ 0.0173} & 0.0543 $\pm$ 0.0300 & 0.0626 $\pm$ 0.0045 & 0.0504 $\pm$ 0.0175 & 0.0723 $\pm$ 0.0410 & 1.2184 $\pm$ 0.1304 \\ \bottomrule
\end{tabular}%
}
\end{table*}

In this section, we describe our experiment setup and analyze the results. We compare our method with strong baselines on three simulated continuous control tasks. We also evaluate EIL on real-world robots.  We aim to understand the extraneousness-aware imitation learning problem by answering the following questions: 1) Does EIL as a framework help the agent imitate from visual demonstrations that contain extraneousness? 2) Can the action-conditioned self-supervised representation differentiate between extraneous and task relevant components in the demonstrations? 3) What are the key factors and design choices for EIL?

\subsection{Simulated Control Tasks}

\subsubsection{Setup and Datasets}
\begin{figure}[!h]
\centering
\subfloat[\textit{Reach}]{\includegraphics[width=0.25\linewidth]{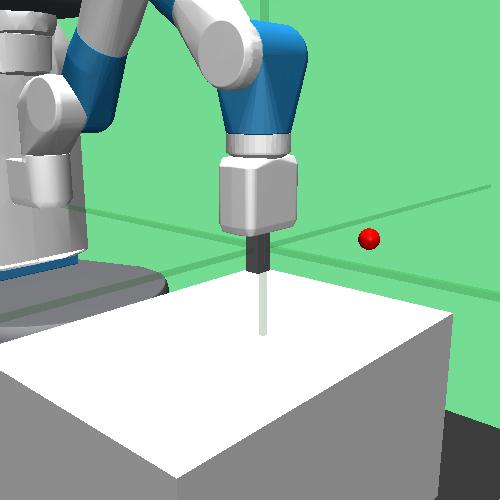}}
\hspace{.2em}
\subfloat[\textit{Push}]{\includegraphics[width=0.25\linewidth]{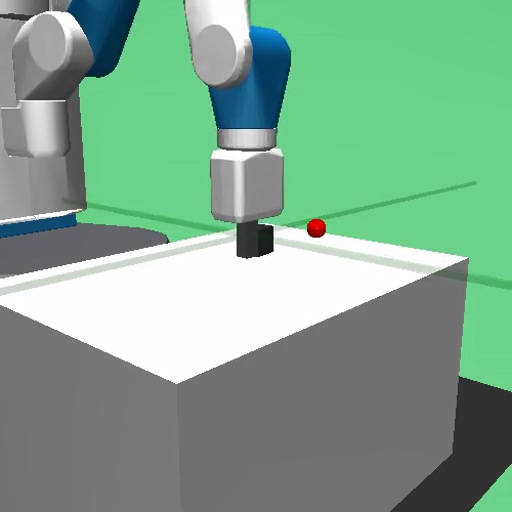}}
\hspace{.2em}
\subfloat[\textit{Stir}]{\includegraphics[width=0.25\linewidth]{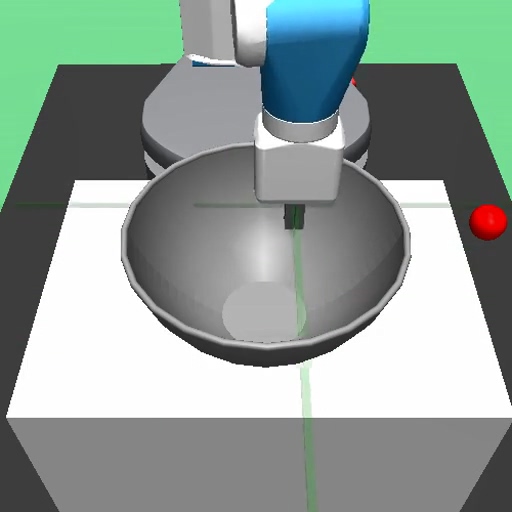}}
\caption{\textbf{The simulated continuous control environments.} \textit{Reach} (a) and \textit{Push} (b) are goal conditioned environments where success is declared when target object is close enough to the destination. In \textit{Stir} (c), success is declared if the end-effector correctly follows a target trajectory. }
\label{fig:simple_navi}
\end{figure}

Visualization and a brief description of the tasks are given in Figure~\ref{fig:simple_navi}. The \textit{Reach} and \textit{Push} tasks' objectives are to move the gripper itself or an object to the target position. For \textit{Stir}, the agent needs to place its gripper inside a bowl, then revolve it around the center of the bowl.
For each task, we collect two datasets -- an extraneous dataset as well as a perfect dataset. In the perfect dataset, all the state-action pairs $\{s_i, a_i\}_{i=1 \cdots N}$ are sampled from an expert policy $\pi_E$. While in the extraneous dataset, there are one or more subsequences that contains $n-m+1$ consecutive steps  $\{o_i, a_i\}_{i=m \cdots n}$ sampled from another policy $\pi_\text{ext}$. 

We note that we do not have any constraints on the non-expert policy $\pi_\text{ext}$, which means it can perform meaningful actions for an irrelevant task or totally random actions. The objects and targets are randomly chosen for every demonstration. For the extraneous parts, we insert locally consistent action sequences at random timesteps.
Specifically, for \textit{Reach} and \textit{Push}, the extraneous action is the agent deviating away from its original trajectory at a random timestep then coming back. For \textit{Stir}, the agent moves the gripper outside the bowl towards a random position and returns, which simulates a real-world scenario where a human fetches an object in the middle of stirring. In the test time, we either provide one perfect reference trajectory or no reference trajectory. 

\subsubsection{Visual Imitation Learning}
For continuous control tasks, we optionally give our policy access to the intrinsic low-level state information of the agent (i.e., position and velocity of the gripper, but not the goal or object). We train a total of 3 random seeds, then choose the one with better performance in validation.
We use the selection scheme from Hussenot et al. for hyperparameter tuning.\ \cite{hussenot2021hyperparameter}.

\subsubsection{Metrics and Evaluation}\label{sec:metric-and-evaluation}
For the goal-conditioned continuous control tasks (\textit{Reach} and \textit{Push}), we use both success rate and minimum goal-object or gripper-object distance as evaluation metrics. For \textit{Stir}, we calculate the mean deviation between the end-effector's trajectory and the target circular orbit, and use this deviation to obtain success rate.
All the experiments are conducted with 3 random seeds. We average over 50 trials for each seed and report the mean and standard deviation.
For all the continuous control tasks, we use the default threshold value defined in the \textit{Reach} and \textit{Push} environment of \SI{0.05}{\metre}.

\subsubsection{Baselines}
We compare EIL with several baselines.

\noindent\textbf{Behavior cloning (BC).} We train a neural network to imitate the mapping from observations to actions through vanilla supervised learning.

\noindent\textbf{Reinforcement learning with embedding-based reward.} We use reinforcement learning to accomplish the task. Instead of the sparse reward of whether the goal is reached, we first obtain a ``goal embedding'' by averaging over the emdeddings of each video's last frame, then set a dense reward of the negative $\ell_2$ distance between the current state-action embedding and the goal embedding. We note that this baseline has the privilege of interacting with the environment.

\noindent \textbf{Time-contrastive networks (TCN)~\cite{sermanet2018time}.} TCN is the predecessor model of TCC, and is originally designed for imitation and reinforcement learning on synchronized multi-view video demonstrations. In our paper, we adopt TCN by substituting the TCC encoder with a TCN network, while the rest of our method (i.e., UVA and visual IL) \mbox{is unchanged.}

\noindent \textbf{Robust imitation learning with co-pseudo-labeling (RIL-Co)~\cite{tangkaratt2020robust}} RIL-Co is the state-of-the-art method for imitation learning from noisy demonstrations. We provide low-level state and action pairs as well as access to the environment to train the policy of RIL-Co. We note that 1) RIL-Co is not a visual imitation learning method and has the advantage of learning from low-level states, and 2) like the RL baseline, RIL-Co requires interaction with the environment.

\noindent \textbf{Random policy.} We sample actions from a uniform random distribution as baseline.

\subsubsection{Experimental Results}
\paragraph{Control Results}

Table~\ref{tab:success-rate} summarizes the mean and standard deviation of success rate on all tasks. The performance of vanilla behavior cloning degrades heavily when extraneousness is present in training data.

Reinforcement learning algorithms struggle to master any of the tasks, even with access to the environment. This may be due to the extraneous frames that cause the task-relevant embeddings to no longer linearly approach the goal embedding. With the continuous tendency of the embedding sequence broken, the distance to the goal embedding may give a wrong guidance, causing the agent to fail.
TCN shows good results in \textit{Reach} and \textit{Push}, but performs poorly in \textit{Stir}. 
\textit{Stir} is different from the other two tasks as it is periodic and could repeat the same states and actions. This violates the assumption of TCN: two frames that are distant in time should have distinct embeddings. 
RIL-Co, despite its privilege of having access to the environment and learning directly from low-level states, can only achieve comparable results in the simplest \textit{Reach} task and cannot perform well in other tasks. A possible reason is the extraneous subsequences violates RIL-Co's assumption for the noise part to be random and inconsistent.
Our method outperforms all the baseline methods and demonstrates its ability to learn from extraneous-rich demonstrations. We also find that EIL outperforms other methods in terms of minimum distance as shown in Table \ref{tab:min-dist}. 

\begin{figure}[!ht]
    \centering
    \includegraphics[width=\columnwidth]{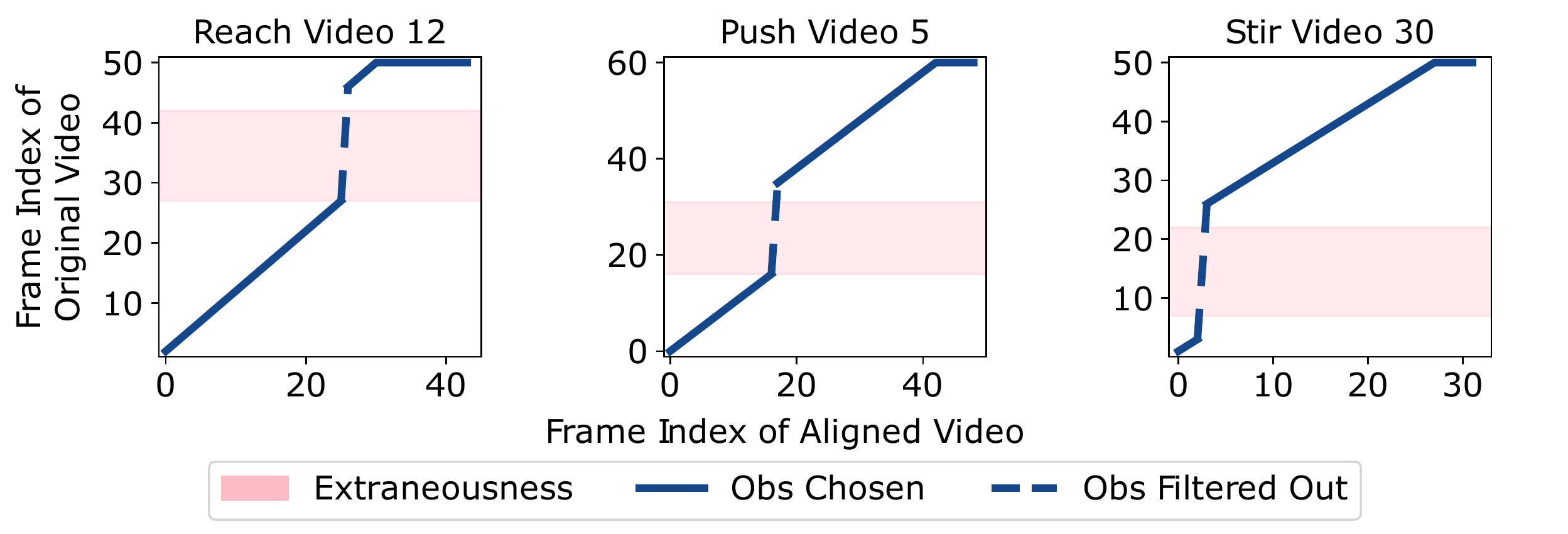}
        \caption{\textbf{UVA's alignment results.} We mark the extraneous subsequence with red shade, observations chosen by UVA with continuous lines, and observations filtered out with dashed lines. Our method successfully skips extraneous part while keeps the others.}
        \label{fig:align_curves}
\end{figure}
\paragraph{UVA Results} 
To evaluate the ability for UVA to exclude task irrelevent frames, we visualize the filter result for each of the task in Figure \ref{fig:align_curves}. Despite some occasional confusion at the border area, UVA can successfully ignore the extraneous parts in the demonstrations. Detailed filtering results could be found at Table~\ref{tab:robot-results} and \ref{tab:ablation-filter}.
\subsection{Real-World Robot Control Tasks}
To further evaluate the performance of EIL and demonstrate the practical value of our method, we adopt a real-world robot arm to learn from demonstrations with extraneousness. We train the arm on the tasks of \textit{Reach} and \textit{Push} using an extraneous dataset with the methods of EIL and behavior cloning.

\subsubsection{Setup and Datasets}
\begin{figure}[h]
    \centering
    \includegraphics[width=\columnwidth]{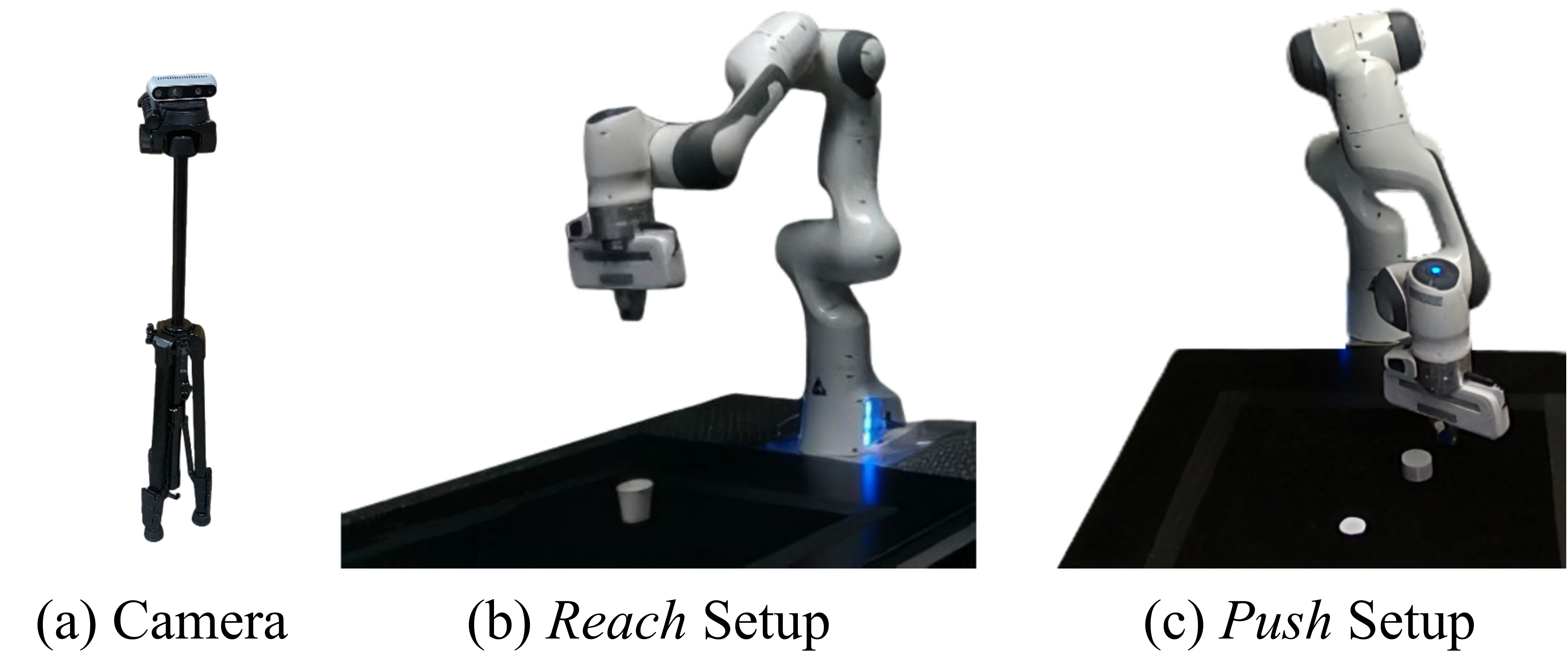}
    \caption{\textbf{Real-World Robot Setup}. The RealSense D435i (a) captures RGB images. The target for \textit{Reach} (b) is a paper cup, while \textit{Push} (c) uses a cylinder as its object and white circle as target.}
    \label{fig:real-robot}
\end{figure}

We use a Franka Emika Panda robot arm as shown in Figure~\ref{fig:real-robot}. The RGB images are captured by an Intel RealSense D435i depth camera.

Similar to the simulator, we collect an extraneous dataset and a perfect dataset. The perfect dataset contains state-action pairs sampled purely from the export policy $\pi_E$, while the extraneous dataset contains one or more subsequences of state-action pairs sampled from another policy $\pi_\text{ext}$. In practice, the camera takes a picture of the current arm, object (in \textit{Push}), and target. A human expert then controls the arm to take the correct action, generating a state-action pair and leading to the next state. Due to this process's costly nature, less  demonstrations (20 videos of 70--100 frames) are collected, and the action space is changed from continuous  to discrete~(front, back, up, down, left, right).

\begin{figure*}[!h]
    \centering
    \includegraphics[width=0.9\textwidth]{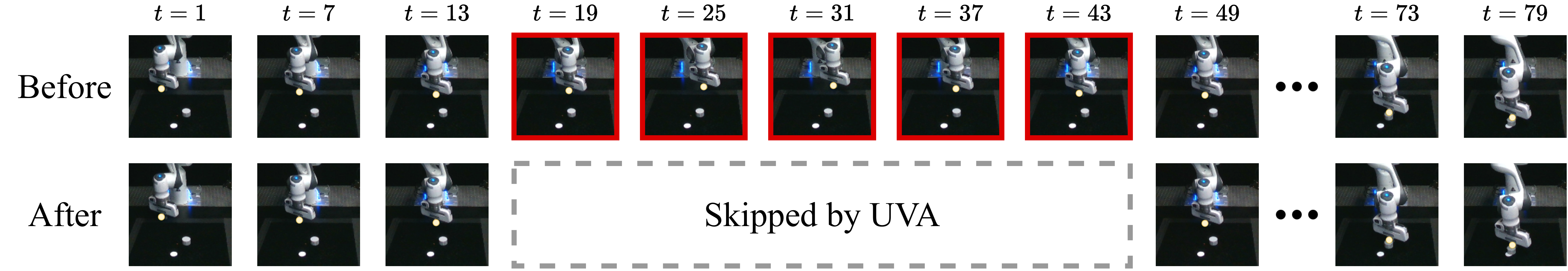}
    \caption{\textbf{Visualized result of filtered demonstration.} The extraneous contents (heading to the other corner in the mid-way) in \textit{Push} are successfully filtered out by UVA. Gripper tips are marked by yellow dots in this graph for clarity.}
    \label{fig:uva-res}
\end{figure*}

\subsubsection{Visual Imitation Learning and Evaluation}

The visual imitation learning process of the real-world robot arm is identical to that of the simulator. However, data augmentation methods of random resizing and cropping are used to overcome jitters of camera positions and illumination conditions.

For evaluation, we measure the distances between the target and the gripper (\textit{Reach}) or object (\textit{Push}). There is also a timestep limit set for every evaluation trajectory. If the agent cannot reach a success state within the time limit, the trajectory will not be considered as successful. Distances are measured from the gripper or object to the center of the target, and the threshold distance of success is the diameter of the target. We evaluate success rate over 20 trials.

\begin{table}[h]
\centering
\caption{Extraneous Percentage (lower is better) and Success Rates (higher is better) for Real-World Robot Learning.}
\label{tab:robot-results}
\begin{tabular}{cccc}
\toprule
\textbf{Task} & \textbf{Metric} & \textbf{BC} & \textbf{EIL (Ours)} \\ \midrule
\multirow{2}{*}{\textit{Reach}} & Extraneous \% & 34.6\% & \textbf{9.6\%} \\
 & Success Rate & 40\% $\pm$ 11\% & \textbf{80\% $\pm$ 9\%} \\ \midrule
\multirow{2}{*}{\textit{Push}} & Extraneous \% & 39.4\% & \textbf{4.1\%} \\
 & Success Rate & 10\% $\pm$ 7\% & \textbf{75\% $\pm$ 10\%} \\ \bottomrule
\end{tabular}
\end{table}

\subsubsection{Experimental Results}
Table~\ref{tab:robot-results} give the results of our real-world robot arm learning experiment. Our method significantly reduces the extraneous ratio in both datasets and improves success rate for imitation learning.

In \textit{Reach}, the BC agent often gets stuck in a local area due to the extraneous segments. For \textit{Push}, failures of BC are mainly due to two reasons: the gripper goes down too much that it hits the table surface, or the object slides out of the gripper's control. The EIL agent, on the other hand, overcomes these difficulties smoothly with the help of UVA.

\subsection{Visualized Results}
We visualize an extraneousness-rich demonstration trajectory and the extracted subsequences with EIL in Figure \ref{fig:uva-res}. From the visualization, we see that the extraneous subsequence is filtered out, and the new demonstration has higher quality, leading to better train results.

\subsection{Ablation Study}
We conduct ablation study on our representation learning and the unsupervised alignment parts of EIL.

\subsubsection{Representation Learning}
EIL learns action-conditioned embeddings from the image encoder $\psi_I$ and action encoder $\psi_A$. For continuous control tasks, intrinsic states 
are also available since the agent has information of itself. Therefore, we may add an intrinsic state encoder $\psi_S$ in addition to $\psi_I$ and $\psi_A$ to provide this extra information. Here we ablate the effect of adding intrinsic state on task success rate and the quality of extraneous frame filtering.

\paragraph{Effects on success rates}
As shown in Table \ref{tab:ablation-success}, EIL is able to obtain the highest success rate in most of the cases. Meanwhile, adding $\psi_S$ \textit{does not} help most of the times. Table~\ref{tab:ablation-success} also infers that adding states during IL training \textit{does not} guarantee improvement in final performance.

\begin{table}[h]
\centering
\caption{Success rates for different encoder configurations to obtain temporal embeddings. Higher is better.}
\label{tab:ablation-success}
\resizebox{\columnwidth}{!}{%
\begin{tabular}{ccccccc}
\toprule
\textbf{Task} & \textbf{IL trained} & \textbf{BC} & \textbf{EIL( $\psi_I$)} & \textbf{EIL($\psi_{I,A}$)} & \textbf{EIL($\psi_{I,A,S}$)} & \textbf{Oracle} \\ \midrule
\multirow{2}{*}{\textit{Reach}} & w/ state & 77\% & \textbf{85\%} & \textbf{85\%} & 80\% & 99\% \\
 & w/o state & 74\% & 70\% & 78\% & \textbf{88\%} & 100\% \\ 
\multirow{2}{*}{\textit{Push}} & w/ state & 66\% & 67\% & \textbf{77\%} & 74\% & 97\% \\
 & w/o state & 57\% & 76\% & \textbf{79\%} & 73\% & 100\% \\ 
\multirow{2}{*}{\textit{Stir}} & w/ state & 55\% & 47\% & 74\% & \textbf{75\%} & 100\% \\
 & w/o state & 54\% & 38\% & \textbf{83\%} & 72\% & 100\% \\ \bottomrule
\end{tabular}%
}
\end{table}

\paragraph{Effects on extraneous filtration}
Table \ref{tab:ablation-filter} shows the percentage of extraneous content in the datasets before and after filtering. UVA is able to decrease extraneous percentage from over 25\% to around 5\% in every task. 

\begin{table}[h]
\centering
\caption{Percentage of extraneous content before and after filtering by different encoder configurations. Lower is better.}
\label{tab:ablation-filter}
\resizebox{\columnwidth}{!}{%
\begin{tabular}{crcccc}
\toprule
\textbf{Task} & \textbf{Quantity} & \textbf{Original} & \textbf{EIL($\psi_I$)} & \textbf{EIL($\psi_{I,A}$)} & \textbf{EIL($\psi_{I, A, S}$)} \\ \midrule
\multirow{3}{*}{\textit{Reach}} & Extraneous & 624 & 108 & 137 & \textbf{91} \\
 & Total & 2200 & 1650 & \textbf{2001} & 1299 \\
 & Extraneous \% & 28.4\% & \textbf{6.5}\% & 6.8\% & 7.0\% \\ \midrule
\multirow{3}{*}{\textit{Push}} & Extraneous & 624 & \textbf{58} & 63 & 144 \\
 & Total & 2489 & 2245 & 2245 & \textbf{2323} \\
 & Extraneous \% & 25.1\% &\textbf{2.6}\% & 2.8\% & 6.2\% \\ \midrule
\multirow{3}{*}{\textit{Stir}} & Extraneous & 624 & 34 & 48 & \textbf{17} \\
 & Total & 2040 & 1221 & \textbf{1338} & 987 \\
 & Extraneous \% & 30.6\% & 2.8\% & 3.6\% & \textbf{1.7}\% \\ \bottomrule
\end{tabular}%
}
\end{table}

\subsubsection{UVA} In this paper, we mainly consider the setting where all the data are not perfect. However, in certain cases, a small amount of perfect demonstrations are also available. We compare the performance of EIL with and without perfect reference trajectory in Table \ref{tab:ablation-dtw}. We use two alignment methods with perfect reference demonstrations: dynamic time warping (DTW) \cite{bellman1959dtw}, and nearest neighbor matching (NN). Nearest neighbor matching directly maps every frame of the perfect demonstration to its nearest neighbor in other videos. DTW ensures a smoother match curve that is chronological. We notice that UVA greatly outperforms DTW and NN despite not having access to perfect data. DTW performs even worse than vanilla Nearest Neighbor (NN) since the locally consistent noise segment disturbs the algorithm and often makes it stick to the segment once it steps into the extraneous area.

\begin{table}[h]
\centering
\caption{Percentage of extraneous content in demonstrations before and after filtering, with or without perfect reference demonstrations. Lower is better.}
\label{tab:ablation-dtw}
\begin{tabular}{cccccc}
\toprule
\textbf{Task} & \multicolumn{1}{c}{\textbf{Quantity}} & \textbf{Original} & \textbf{UVA} & \textbf{DTW} & \textbf{NN} \\ \midrule
\multirow{3}{*}{\textit{Reach}} & Extraneous & 624 & 137 & 237 & 152 \\
 & Total & 2200 & 2001 & 2040 & 2040 \\
 & Extraneous \% & 28.4\% & \textbf{6.8\%} & 11.6\% & 7.5\% \\\midrule
\multirow{3}{*}{\textit{Push}} & Extraneous & 624 & 63 & 180 & 98 \\
 & Total & 2489 & 2245 & 2440 & 2440 \\
 & Extraneous \% & 25.1\% & \textbf{2.8\%} & 7.4\% & 4.0\% \\\midrule
\multirow{3}{*}{\textit{Stir}} & Extraneous & 624 & 48 & 357 & 195 \\
 & Total & 2040 & 1338 & 2040 & 2040 \\
 & Extraneous \% & 30.6\% & \textbf{3.6\%} & 17.5\% & 9.6\% \\ \bottomrule
\end{tabular}%
\end{table}

\section{Conclusions and Future Work}
This paper focuses on visual imitation from demonstrations where temporally consistent yet task-irrelevant subsequences are present. We propose Extraneousness-Aware Imitation Learning (EIL), a framework that enables agents to identify extraneous subsequences from visual demonstrations via self-supervised learning. 
%
Empirical results show that EIL outperforms strong baselines on continuous and discrete control benchmarks in both simulator and the real-world. An exciting direction for future work is to integrate EIL with robotic learning from human demonstration methods, which usually deals with videos containing rich task-irrelevant segments. Another extension would be scenarios where observations are vulnerable to temporary corruption, like the dazzling light in a driving scene. EIL could filter out these corrupted observations as well since they are inconsistent with the normal observation sequence.

\bibliography{eil}
\bibliographystyle{IEEEtran}

\end{document}